%% file: main.tex
\documentclass{article}
\usepackage[preprint]{neurips_2021}

\usepackage[utf8]{inputenc} 
\usepackage[T1]{fontenc}    
\usepackage[colorlinks,citecolor=blue]{hyperref}
\usepackage{url}            
\usepackage{booktabs}       
\usepackage{amsfonts}       
\usepackage{nicefrac}       
\usepackage{microtype}      
\usepackage{xcolor}         

\include{commands}

\usepackage{amssymb}
\usepackage{amsmath}
\usepackage{wrapfig,lipsum}

\newtheorem{thm}{Theorem}
\newtheorem{lem}{Lemma}

\newtheorem{defi}{Definition}

\newcommand{\afgcl}{\textsc{AF-GCL}}

\usepackage{floatrow}
\newfloatcommand{capbtabbox}{table}[][\FBwidth]
\usepackage{blindtext}

\title{Augmentation-Free Graph Contrastive Learning with Performance Guarantee}

\author{%
  Haonan Wang\\
  University of Illinois Urbana-Champaign\\
  \texttt{haonan3@illinois.edu} \\
   \And
   Jieyu Zhang\\
   University of Washington\\
   \texttt{jieyuz2@cs.washington.edu}\\
 \And
   Qi Zhu\\
  University of Illinois Urbana-Champaign\\
  \texttt{qiz3@illinois.edu} \\
 \And
   Wei Huang \thanks{Corresponding Author} \\
  University of Technology Sydney\\
  \texttt{weihuang.uts@gmail.com} \\
}

\begin{document}
\maketitle
\input{sec-abstract}
\input{sec-intro}
\input{sec-related}
\input{sec-preliminary}
\input{sec-analysis}
\input{sec-method}

\input{sec-theory}
\input{sec-exp}
\input{sec-conclusion}


\clearpage
\input{sec-appendix}

\end{document}

%% file: commands.tex
\usepackage{booktabs} 
\usepackage{amsmath,url}

\usepackage{amssymb}
 
\usepackage{amsfonts}
\usepackage{multirow}
\usepackage[ruled,vlined]{algorithm2e}
\usepackage{algorithmic,amsmath}
\usepackage{color}
\usepackage{enumitem}
\usepackage{dsfont}
\usepackage{graphicx}
\usepackage{subcaption}
\usepackage{bbm}

\newcommand{\mylistbegin}{
  \begin{list}{$\bullet$}
   {
     \setlength{\itemsep}{-2pt}
     \setlength{\leftmargin}{1em}
     \setlength{\labelwidth}{1em}
     \setlength{\labelsep}{0.5em} } }
\newcommand{\mylistend}{
   \end{list}  }

\newcommand{\ie}{\textit{i.e.}}

\newcommand{\ceil}[1]{\lceil #1 \rceil}

\DeclareMathOperator*{\argmax}{arg\,max}

%% file: sec-abstract.tex
\begin{abstract}
Graph contrastive learning (GCL) is the most representative and prevalent self-supervised learning approach for graph-structured data. Despite its remarkable success, existing GCL methods highly rely on an augmentation scheme to learn the representations invariant across different augmentation views.
In this work, we revisit such a convention in GCL through examining the effect of augmentation techniques on graph data via the lens of spectral theory.
We found that graph augmentations preserve the low-frequency components and perturb the middle- and high-frequency components of the graph, 
which contributes to the success of GCL algorithms on homophilic graphs but hinder its application on heterophilic graphs, due to the high-frequency preference of heterophilic data.
Motivated by this, we propose a novel, theoretically-principled, and augmentation-free GCL method, named \afgcl, that (1) leverages the features aggregated by Graph Neural Network to construct the self-supervision signal instead of augmentations and therefore (2) is less sensitive to the graph homophily degree.
Theoretically, We present the performance guarantee for \afgcl\ as well as an analysis for understanding the efficacy of \afgcl.
Extensive experiments on 14 benchmark datasets with varying degrees of heterophily show that \afgcl\  presents competitive or better performance on homophilic graphs and outperforms all existing state-of-the-art GCL methods on heterophilic graphs with significantly less computational overhead.
\end{abstract}

%% file: sec-intro.tex
\section{Introduction}
\label{intro}
Graph Neural Networks (GNNs)~\cite{kipf2016semi, xu2018powerful, velivckovic2017graph, hamilton2017inductive} have attracted great attention due to their success in various applications involving graph-structured data, such as node classification~\cite{kipf2016semi}, edge prediction~\cite{kipf2016variational}, graph classification~\cite{xu2018powerful}, etc. Most of these tasks are semi-supervised and therefore require a certain number of labels to guide the learning process. 
However, in many real-world applications (e.g., chemistry and healthcare), labels are scarcely available.
Self-supervised learning (SSL), as an appropriate paradigm for such label-scarce settings, has been extensively studied in Computer Vision (CV). Besides, contrastive learning, the most representative SSL technique, has achieved state-of-the-art performance~\cite{jaiswal2021survey}.
This has motivated the self-supervised learning, especially contrastive learning, approaches~\cite{xie2021self} on graph data.

Contrastive learning is essentially learning representations invariant to data augmentations which are thoroughly explored on visual data~\cite{hataya2020faster, lim2019fast}.
Leveraging the same paradigm, Graph Contrastive Learning (GCL) encourages the representations to contain as less information about the way the inputs are transformed as possible during training, i.e. to be invariant to a set of manually specified transformations. 
However, the irregular structure of the graph complicates the adaptation of augmentation techniques used on images, and prevents the extending of theoretical analysis for visual contrastive learning to a graph setting.
In recent years, many works~\cite{you2020graph, peng2020graph, hassani2020contrastive, zhu2021graph, zhu2020deep, zhu2020transfer} focus on the empirical design of hand-craft graph augmentations for graph contrastive learning from various levels, including node dropping, edge perturbation, attribute masking, subgraph~\cite{you2020graph}, and graph diffusion~\cite{hassani2020contrastive}.
Although experiments have demonstrated the effectiveness of the GCL algorithms~\cite{zhu2021empirical}, those empirical studies are limited to the homophilic graphs, 
where the linked nodes are likely from the same class, e.g. social network and citation networks~\cite{mcpherson2001birds}.
In heterophilic graphs, similar nodes are often far apart (e.g., the majority of people tend to connect with people of the opposite gender~\cite{zhu2020beyond} in dating networks), which urges the investigation on the generalization of GCL frameworks on both homophilic and heterophilic graphs.


To fill this gap, we first investigate the empirical success of GCL algorithms. As discussed in~\cite{you2020graph, zhu2021empirical}, existing GCL algorithms learn invariant representations across different graph augmentation views.
In this work, we take a closer look at what information is preserved or perturbed by commonly used graph augmentations.
By analyzing the graph structure and features from the view of frequency,
we observe that graph augmentations mainly preserve low-frequency information and corrupt middle and high-frequency information of the graph. 
By enforcing the model to learn representations invariant to the perturbations through maximizing the agreement between a graph and its augmentations, the learned representations will only retain the low-frequency information.
As demonstrated in~\cite{balcilar2020analyzing}, the low-frequency information is critical for the homophilic graph. However, for heterophilic graph, the low-frequency information is insufficient for learning effective representations. Under this circumstance, the middle and high-frequency information, capturing the difference between nodes, may be more effective~\cite{bo2021beyond, li2021beyond}, but typically overlooked by existing GCL algorithms.
Thus, it rises a natural question, that is, \emph{is it possible to design a generic graph contrastive learning method effective on both homophilic and heterophilic graphs?}

In this work, we answer the above question affirmatively by providing a new perspective of achieving SSL on graphs.
Specifically, based on our analysis of the concentration property of aggregated features on both homophilic and heterophilic graphs, we propose a novel {\it augmentation-free graph contrastive learning} method, named \afgcl.
Different from the large body of previous works on the graph contrastive learning, in which the construction of self-supervision signals is heavily relying on graph augmentation, 
\afgcl\ constructs positive/negative pairs based on the aggregated node features and directly optimizes the distance of representations in high-dimensional space.
As a simple yet effective approach, \afgcl\ frees the model from dual branch design and graph augmentations, which enable the proposed method to easily scale to large graphs.

In addition, we present a theoretical guarantee for the performance of embedding learned by AF-GCL in downstream tasks,  and our theoretical analysis provides an understanding of when and why \afgcl\ can work well.
Experimental results show that \afgcl\ can outperform or be competitive with state-of-the-art GCL algorithms on 14 graph benchmarks datasets with varying degrees of heterophily.
Furthermore, we analyze the computational complexity of \afgcl\ and empirically show that our method performs well with significantly less computational overhead.
Our contribution could be summarized as:
\begin{itemize}[leftmargin=15pt]
\vspace{-10pt}
\setlength\itemsep{-0.2em}
\item We first analyze the efficacy of graph augmentation techniques for GCL as well as its limitations from a spectral point of view. We show that augmentation-based GCL is sensitive to the graph's homophily degree.
\item We then illustrate the concentration property of representations obtained by the neighborhood feature aggregation, which in turn inspires our novel augmentation-free graph contrastive learning method, \afgcl.
\item We further present a theoretical guarantee for the performance of \afgcl, as well as the analyses of \afgcl's robustness to the graph's homophily degree.
\item Experimental results show that without complex designs, compared with SOTA GCL methods, \afgcl\ achieves competitive or better performance on 8 homophilic graph benchmarks and 6 heterophilic graph benchmarks, with significantly less computational overhead.
\end{itemize}

%% file: sec-related.tex
\section{Related Work}
\label{sec:Related Work}
\vspace{-8pt}
\subsection{Graph Contrastive Learning}
\vspace{-5pt}
Contrastive learning aims to learn consistent representations under proper transformations and has been widely applied to the visual domain.
Graph Contrastive Learning (GCL) leverages the idea of CL on the graph data. However, due to
the complex, irregular structure of graph data, it is more challenging to design appropriate strategies for constructing positive and negative samples on the graph than that on visual or textual data.
Regarding graph augmentation, many previous studies~\cite{you2020graph, peng2020graph, hassani2020contrastive, zhu2021empirical, zhu2021graph, zhu2020deep, zhu2020transfer} propose data augmentation techniques for general graph-structured data, e.g., attribute masking, edge removing, edge adding, subgraph, graph diffusion.
Specifically, MVGRL~\cite{hassani2020contrastive} employs graph diffusion to generate graph views with more global information; 
GCC~\cite{qiu2020gcc} use the subgraph induced by random walks to serve as different views.
GraphCL~\cite{you2020graph} study multiple augmentation methods for graph-level representation learning.
GRACE~\cite{zhu2020deep} constructs node-node pairs by using edge removing and feature masking.
GCA~\cite{zhu2021graph} proposes adaptive augmentation techniques to further consider important topology and attribute information. 
BGRL~\cite{grill2020bootstrap} gets rid of the design of negative pairs, but its design of positive pairs also relies on edge removing and feature masking.
We summarized the graph augmentation methods employed by the representative GCL methods in Table~\ref{tab:graph_aug_summary}.
To the best of our knowledge, the current state-of-the-art GCL algorithms are highly reliant on graph augmentations, but none of the existing work studies the effect and limitation of current graph augmentation techniques in GCL.

\begin{table*}[h]
\vspace{-2pt}
\centering
\caption{Summary of graph augmentations used by representative GCL models. $\text{Multiple}^*$ denotes multiple augmentation methods including edge removing, edge adding, node dropping and subgraph induced by random walks.}
\label{tab:graph_aug_summary}
\begin{tabular}{c|c|c}
\toprule
\text { Method }  & \text { Topology Aug. } & \text { Feature Aug. } \\
\hline 
\text { MVGRL~\cite{hassani2020contrastive} } & \text { Diffusion } & - \\
\text { GCC~\cite{qiu2020gcc}} & \text { Subgraph } & - \\
\text { GraphCL~\cite{you2020graph} } & \text { $\text{Multiple}^*$ } & \text { Feature Dropout  } \\
\text { GRACE~\cite{zhu2020deep} }  & \text { Edge Removing } & \text { Feature Masking } \\
\text { GCA~\cite{zhu2021graph} }  & \text { Edge Removing } & \text { Feature Masking } \\
\text { BGRL~\cite{grill2020bootstrap} }  & \text { Edge Removing } & \text { Feature Masking } \\
\bottomrule
\end{tabular}
\vspace{-2pt}
\end{table*}

\subsection{Understanding Contrastive Learning}
\vspace{-8pt}
Previous theoretical guarantees for contrastive learning follow conditional independence
assumption (or its variants)~\cite{arora2019theoretical, lee2020predicting, tosh2021contrastive, tsai2020self}. 
Specifically, they assume the two contrastive views are independent conditioned on the label
and show that contrastive learning can provably learn representations beneficial for downstream
tasks. 
In addition, Wang et al.\cite{wang2020understanding} investigated the representation geometry of supervised contrastive loss and showed that the contrastive loss favors data representation uniformly distributed over the unit sphere yet aligning across semantically similar samples.
Haochen et al.\cite{haochen2021provable} analyzed the contrastive learning on the augmented image dataset through the novel concept augmentation graph with a new loss function that performs spectral decomposition on the graph. 
However, all those theoretical analyses mainly focus on the classification problem with image datasets.
Since graphs are far more complex due to the non-Euclidean property, the analysis for image classification cannot be trivially extended to graph setting.

Besides, on the other line of research,
contrastive learning methods~\cite{khosla2020supervised, oord2018representation, peng2020graph, zhu2021graph} leveraging the information maximization principle (InfoMax)~\cite{linsker1988infomaxprinciple} aim to maximize the Mutual Information (MI) between the representation of one data point and that of its augmented version by contrasting positive pairs with negative-sampled counterparts. 
The key idea is that maximizing mutual information between representations extracted from multiple views can force the representations to capture information about higher-level factors (e.g., presence of certain objects or occurrence of certain events) that broadly affect the shared context.
The employed loss functions, e.g. Information Noise Contrastive Estimation (InfoNCE) and Jensen-Shannon Divergence (JSD),  are proved to be lower bounds of MI~\cite{gutmann2010noise, nowozin2016f, poole2019variational}.
Although the idea of information maximization principle has been used in GCL domain~\cite{velickovic2019deep, zhu2020transfer, zhu2021graph}, the higher-level factors invariant across different graph augmentation views is under-defined.
In this work, we take a closer look at the graph augmentations via the lens of spectral theory and analyze what information is preserved in different augmentation views.

%% file: sec-preliminary.tex
\section{Preliminary}
\label{sec:preliminary}

\subsection{Notation}
\label{subsec:notation}
Let $\mathcal{G}=(\mathcal{V}, \mathcal{E})$ denote an undirected graph, where $\mathcal{V} = \{v_i\}_{i\in[N]}$ and $\mathcal{E} \subseteq$ $\mathcal{V} \times \mathcal{V}$ denote the node set and the edge set respectively.
We denote the number of nodes and edges as $N$ and $E$, and the label of nodes as $\mathbf{y}\in \mathbb{R}^N$, in which $y_i \in [1, c], c\geq2$ is the number of classes.
The associated node feature matrix denotes as $\mathbf{X} \in \mathbb{R}^{N \times F}$, where $\mathbf{x}_{i} \in \mathbb{R}^{F}$ is the feature of node $v_i \in \mathcal{V}$ and $F$ is the input feature dimension.
The adjacent matrix denotes as $\mathbf{A} \in\{0,1\}^{N \times N}$, where $\mathbf{A}_{i j}=1$ if $\left(v_i, v_j\right) \in \mathcal{E}$.
Our objective is to unsupervisedly learn a GNN encoder $f_{\theta}: \mathbf{X}, \mathbf{A} \rightarrow \mathbb{R}^{N \times K}$ receiving the node features and graph structure as input, that produces node representations in low dimensionality, i.e., $K \ll F$. The representations can benefit the downstream supervised or semi-supervised tasks, e.g., node classification. 

\subsection{Homophilic and Heterophilic Graph}
Various metrics have been proposed to measure the homophily degree of a graph.
Here we adopt two representative metrics, namely, node homophily and edge homophily. The edge homophily~\cite{zhu2020beyond} is the proportion of edges that connect two nodes of the same class:
\begin{equation}
\begin{small}
\begin{aligned}
h_{edge}=\frac{\left|\left\{\left(v_{i}, v_{j}\right):\left(v_{i}, v_{j}\right) \in \mathcal{E} \wedge y_{i}=y_{j}\right\}\right|}{E},
\end{aligned}
\end{small}
\end{equation}
And the node homophily~\cite{pei2020geom} is defined as, 
\begin{equation}
\begin{small}
\begin{aligned}
h_{node}=\frac{1}{N} \sum_{v_i\in \mathcal{V}} \frac{\left|\left\{v_{j}:\left(v_{i}, v_{j}\right) \in \mathcal{E} \wedge y_{i}=y_{j}\right\}\right|}{ \left| \left\{ v_{j}:\left(v_{i}, v_{j}\right) \in \mathcal{E} \right\} \right| },
\end{aligned}
\end{small}
\end{equation}
which evaluates the average proportion of edge-label consistency of all nodes.
They are all in the range of $[0, 1]$ and a value close to $1$ corresponds to strong homophily while a value close to $0$ indicates strong heterophily.
As conventional, we refer the graph with high homophily degree as homophilic graph, and the graph with low homophily degree as heterophilic graph. And we provided the homophily degree of graph considered in this work in Table~\ref{tab:data_stat}.

\subsection{Graph Laplacian and Variants}
\label{subsec:laplacian_concept}
We define the Laplacian matrix of the graph as $\mathbf{L}=\mathbf{D}-\mathbf{A}$, where $\mathbf{D}=\operatorname{diag}\left(d_{1}, \ldots, d_{N}\right)$, $d_{i}=\sum_{j} \mathbf{A}_{i, j}$. 
The symmetric normalized Laplacian, is defined as $\mathbf{L}_{s y m}=\mathbf{D}^{-\frac{1}{2}} \mathbf{L} \mathbf{D}^{-\frac{1}{2}}=\mathbf{U} \mathbf{\Lambda} \mathbf{U}^{\top}$. Here $\mathbf{U} \in \mathbb{R}^{N \times N}=\left[\mathbf{u}_{1}, \ldots, \mathbf{u}_{N}\right]$, where $\mathbf{u}_{i} \in \mathbb{R}^{N}$ denotes the $i$-th eigenvector of $\mathbf{L}_{s y m}$ and $\mathbf{\Lambda}=$ $\operatorname{diag}\left(\lambda_{1}, \ldots, \lambda_{N}\right)$ is the corresponding eigenvalue matrix.
$\lambda_{1}$ and $\lambda_{N}$ be the smallest and largest eigenvalue respectively.
The affinity (transition) matrices can be derived from the Laplacian matrix, $\mathbf{A}_{\text {sym }}=\mathbf{I}-\mathbf{L}_{\text {sym }}=\mathbf{D}^{-1 / 2} \mathbf{A} \mathbf{D}^{-1 / 2}=\mathbf{U} (\mathbf{I} - \mathbf{\Lambda})\mathbf{U}^{\top}$.
%
%
The $\mathbf{L}_{sym}$ has eigenvalue from $0$ to $2$ and is widely used in the design of spectral graph neural networks, such as Graph Convolutional Network (GCN)~\cite{kipf2016semi}.

For the Laplacian matrix, the smaller eigenvalue is corresponding to the lower frequency~\cite{balcilar2020analyzing}.
Following the previous work~\cite{entezari2020all}, we define the decomposed components of ${\mathbf{L}}_{sym}$ under different frequency bands as ${\mathbf{L}}^m_{sym}$ which has eigenvalues in $\big[{\lambda}_{N}\cdot\frac{(m-1)}{M}, {\lambda}_{N}\cdot\frac{m}{M}\big)$, and $m \in [1, M], M\in \mathcal{Z}^+$ denotes the number of partition of the spectrum. 
More specifically, 
$
{\mathbf{L}}^m_{s y m}={\mathbf{U}} {\mathbf{\Lambda}}^m {\mathbf{U}}^{\top}, {\mathbf{\Lambda}}^m = \operatorname{diag}\left({\lambda}_{1}^m, \ldots, {\lambda}_{N}^m\right)$, where for $i \in [1, N]$,
$$
{\lambda}_{i}^m= \begin{cases} {\lambda}_{i}, \text { if } \lambda_{i} \in \big[{\lambda}_{N}\cdot\frac{(m-1)}{M},  {\lambda}_{N}\cdot\frac{m}{M}\big) \\
0, \text { otherwise }\end{cases},
$$
Note, the sum of all decomposed components is equal to the symmetric normalized Laplacian matrix, ${\mathbf{L}}_{sym} = \sum_{m=0}^{\ceil{N/M}}{\mathbf{L}}^m_{sym}$.

%% file: sec-analysis.tex
\section{Revisiting Graph Augmentations}
\label{sec:aug_study}
Graph Contrastive Learning (GCL) aims to learn representations that are invariant to different augmentations. However, it is rarely studied what information is preserved or perturbed across augmented graph views. In this section, we attempt to identify such information by examining the effect of graph augmentation techniques on graph geometric structure and node attributes from the spectral perspective.

\subsection{Representative Augmentation Techniques}
According to Table~\ref{tab:graph_aug_summary}, the most commonly used four graph augmentation techniques for GCL are: attribute masking, edge adding, edge dropping~\cite{you2020graph} and graph diffusion~\cite{you2020graph}.
\begin{itemize}
  \item Attribute Masking: randomly masks a fraction of features for all the nodes.
  \item Edge Adding/Dropping: randomly adds/drops a fraction of edges from the original graph.
  \item Graph Diffusion: the Personalized PageRank (PPR) based graph diffusion is defined as, $\alpha\left(\mathbf{I}-(1-\alpha) \mathbf{A}_{sym}\right)^{-1}$, where $\alpha$ is the diffusion factor.
\end{itemize}
\subsection{Effect of Augmentations on Geometric Structure}
\begin{wrapfigure}{r}{6cm}
\vspace{-25pt}
  \begin{center}
    \includegraphics[width=6.0cm]{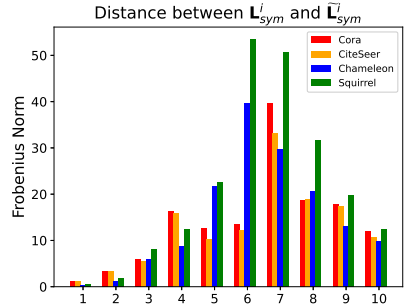}
  \end{center}
  \vspace{-10pt}
  \caption{{Frobenius distance between the decomposed symmetric normalized Laplacian matrices of the original graph and the augmented graph with 20\% edge dropping.} The experiment is independent conducted 10 times and the average value is reported.}
  \label{fig:er_fm_observation}
\vspace{-10pt}
\end{wrapfigure}
First, we investigate the effect of different augmentations, e.g., edge adding, edge dropping and graph diffusion, on adjacency matrix.
As we introduced in Section~\ref{subsec:laplacian_concept}, the graph Laplacians are widely used in GNNs, therefore we measure the difference of Laplacians caused by graph augmentations in different frequency bands.
The $m$-th component symmetric normalized Laplacian is defined as ${\mathbf{L}}^{m}_{sym}$. Correspondingly, we denote the decomposed $m$-th symmetric normalized Laplacian for the augmented graph as ${\widetilde{\mathbf{L}}}_{sym}^{m}$. 
To measure the impact of graph augmentations for different frequency components, we employ the Frobenius norm as the metric to measure the distance, $\|{\mathbf{L}}^{m}_{sym} - {\widetilde{\mathbf{L}}}^{m}_{sym}\|_F$. 
The results of edge dropping on two homophilic graphs, e.g., Cora and CiteSeer~\cite{sen2008collective, namata2012query}, and two heterophilic graphs, e.g., Chameleon and Squirrel~\cite{rozemberczki2021twitch}, are summarized in Figure~\ref{fig:er_fm_observation} and the results of other graph augmentation techniques are in Appendix~\ref{apd:aug_study}. We observed that graph augmentations have less impact on low-frequency components and more impact on middle and high-frequency components. 
Our conclusion is aligned with the previous works~\cite{entezari2020all, chang2021not} in the graph adversarial attack domain, in which they find that, for the graph structure modification, perturbations on the low-frequency components are smaller than that in the middle or high-frequency ones.

\begin{wraptable}{r}{5cm}
\vspace{-10pt}
\caption{Distance between original node features and augmented node features with 30\% attribute dropping. We set $R=0.8\times F$.}
\label{tab:feat_freq_analysis}
\vspace{-10pt}
\begin{tabular}{ccc}\\
\toprule
 & F-Low & F-High \\\hline
Cora & 0 & 7.949 \\  \hline
CiteSeer & 0 & 5.609 \\  \hline
Chameleon & 0 & 12.566 \\  \hline
Squirrel & 0 & 15.836 \\  
\bottomrule
\end{tabular}
\vspace{-20pt}
\end{wraptable} 

\subsection{Effect of Augmentations on Features}
To further study the effect of the commonly used graph augmentation method, attribute masking, on node attribute from spectral view. We denote the Fourier transform and inverse Fourier transform as $\mathcal{F}(\cdot)$ and $\mathcal{F}^{-1}(\cdot)$.
We use $\mathbf{H}^\mathcal{F}$ to denote the transformed node features. Therefore, we have $\mathbf{H}^{\mathcal{F}}=\mathcal{F}(\mathbf{X})$ and $\mathbf{X}=\mathcal{F}^{-1}(\mathbf{H}^{\mathcal{F}})$.
We decompose the node attribute $\mathbf{X}=\left\{\mathbf{X}^{l}, \mathbf{X}^{h}\right\}$, where $\mathbf{X}^{l}$ and $\mathbf{X}^{h}$ denote the low-frequency and high-frequency components of $\mathbf{X}$ respectively. We have the following four equations:
$$
\begin{array}{rlr}
\mathbf{H}^{\mathcal{F}}=\mathcal{F}(\mathbf{X}), & \mathbf{H}^{l}, \mathbf{H}^{h}=t(\mathbf{H}^{\mathcal{F}}; R), \\
\mathbf{X}^{l}={\mathcal{F}}^{-1}\left(\mathbf{H}^{l}\right), & \mathbf{X}^{h}=\mathcal{F}^{-1}\left(\mathbf{H}^{\mathcal{F}}\right),
\end{array}
$$
where $t(\cdot ; R)$ denotes a thresholding function that separates the low and high frequency components from $\mathbf{H}^f$ according to a hyperparameter, $m$. Because the column of $\mathbf{H}^{\mathcal{F}}$ in the left is corresponding to the low frequency component, we define $t(\cdot ; m)$ as:
\begin{equation}
\begin{aligned}
&\mathbf{H}^{l}_{ij}= \begin{cases}\mathbf{H}^{\mathcal{F}}_{ij}, & \text { if } j \leq R \\
0, & \text { otherwise }\end{cases}, 
&\mathbf{H}^{h}_{ij}= \begin{cases}0, & \text { if } j \leq R \\
\mathbf{H}^{\mathcal{F}}_{ij}, & \text { otherwise }\end{cases}.
\end{aligned}
\end{equation}
Further, we denote the node attribute with attribute masking as $\widetilde{\mathbf{X}}$ and its corresponding low and high frequency components as $\widetilde{\mathbf{X}}^l$, $\widetilde{\mathbf{X}}^h$.
We investigate the influence of attribute masking on node features by computing the Frobenius distance of matrix, and denote $\|{\mathbf{X}}^l - \widetilde{\mathbf{X}}^l \|_F$ as F-norm-low
and  $\|{\mathbf{X}}^h - \widetilde{\mathbf{X}}^h \|_F$ as F-norm-high.
The results on four datasets are summrized in Table~\ref{tab:feat_freq_analysis}. We surprisingly find that the attribute masking will always affect the high frequency component.
\subsection{Concluding Remarks}
As demonstrated by previous works~\cite{bo2021beyond, li2021beyond},  for heterophilic graphs, the information carried by high-frequency components is more effective for downstream classification performance.
However, as we analyzed, the middle- and high-frequency information are perturbed by commonly used graph augmentation techniques. With the information maximization objective, only the invariant information (low-frequency) is encouraged to be captured by the learned embedding~\cite{you2020graph}.
Although existing graph augmentation algorithms promote the success of GCL on traditional (homophilic) benchmarks, they result in sub-optimal representations when the high-frequency information is crucial (heterophilic graphs).



%% file: sec-method.tex
\section{Methodology}
As analyzed in the previous section, the aforementioned graph augmentation techniques are less effective for heterophilic graphs (see also experimental results in Table~\ref{tab:heter_result}). To design a universal self-supervision signal, we are motivated to analyze the concentration property of aggregated node feature $\mathbf{Z}$ (Section~\ref{sec:concentration}) for both homophilic and heterophilic graphs. Namely, nodes of the same class are closer with each other under different degree of homophily.
Leveraging the concentration property, we propose an augmentation-free method (Section~\ref{sec:Method}), \afgcl, to construct positive and negative pairs for contrastive learning.

\begin{figure}[h]
\centering
\includegraphics[width=13cm]{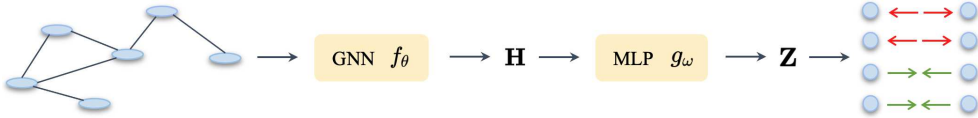}
\caption{Overview of the proposed \afgcl. The input graph is encoded by a graph neural network $f_{\theta}$ and a following projection head $g_{\omega}$. The contrastive pairs are constructed based on the representation $\mathbf{Z}$.}
\label{fig:model}
\end{figure}
\subsection{Analysis of Aggregated Features}
\label{sec:concentration}

\textbf{Assumptions on graph data.}
To obtain analytic and conceptual insights of the aggregated features, we firstly describe the graph data we considered.
We assume that the node 
feature follows the Gaussian mixture model \cite{reynolds2009gaussian}. For simplicity, we focus on the binary classification problem. Conditional on the (binary-) label $y$ and a latent vector $\boldsymbol{\mu} \sim \mathcal{N}( \mathbf{0}, \mathbf{I}_F/F)$ where the identity matrix $\mathbf{I}_F \in \mathbb{R}^{F\times F}$, the features are governed by:
\begin{small}
\begin{equation} \label{eq:feature}
{\bf x}_i = y_i \boldsymbol{\mu} + \frac{ {\mathbf{q}}_i}{\sqrt{F}},
\end{equation}
\end{small}
where random variable $\mathbf{q}_i \in \mathbb{R}^F$ has independent standard normal entries and $y_i \in \{-1, 1 \}$ representing latent classes with abuse of notation. Then, the features of nodes with class $y_i$ follow the same distribution depending on $y_i$, \ie, $\mathbf{x}_i \sim P_{y_i}(\mathbf{x})$.
Furthermore, we make an assumption on the neighborhood patterns,
For node $i$, its neighbor's labels are independently sampled from a distribution $P({y_i})$.

\textbf{Remark.} The above assumption implies that the neighbor's label is generated from a distribution only dependent on the label of the central node, which contains both cases of homophily and heterophily. With this assumption, we prove the following Lemma~\ref{lem:emb} that all aggregated node features with the same label have the same embedding regardless of homophily or heterophily. Specifically, we define learned embedding through a GCN and MLP by $\mathbf{Z}$ as shown in Figure \ref{fig:model}, and $\mathbf{Z}_i$ is the learned embedding with respect to input $\mathbf{x}_i$. To simplify the analysis, we introduce  $\mathbf{W}$ being the equivalent linear weight after dropping non-linearity in GCN and MLP.  

\begin{lem}[Adaption of Theorem 1 in \cite{ma2021homophily}] \label{lem:emb}
Consider a graph $G$ following the graph assumption and Eq. (\ref{eq:feature}), then the expectation of embedding is given by
\begin{equation}
\mathbb{E}[\mathbf{Z}_i] = \mathbf{W} ~ \mathbb{E}_{y \sim P({y_i}), \mathbf{x} \sim P_y(\mathbf{x}) } [\mathbf{x}],
\end{equation}
Furthermore, with probability at least $1-\delta$ over the distribution for graph, we have:
\begin{equation}
\| \mathbf{Z}_i - \mathbb{E}[\mathbf{Z}_i]   \|_2 \le \sqrt{ \frac{\sigma^2_{\max}(\mathbf{W})F  \log(2F/\delta)}{2\mathbf{D}_{ii}\|\mathbf{x} \|_{\psi_2}} },
\end{equation}
where the sub-gaussian norms $\|\mathbf{x} \|_{\psi_2} \equiv \min_i \|\mathbf{x}_{i,d} \|_{\psi_2}$, $d \in [1,F]$ and $\sigma^2_{\max}(\mathbf{W})$ is the largest singular value of $\mathbf{W}$, because each dimension in feature is independently distributed.
\end{lem}

We leave the proof of the above lemma in Appendix~\ref{apd: lemma_emb_proof}.
The above lemma indicates that, for any graph where the feature and neighborhood pattern of each node is sampled from the distributions depending on the node label, the GCN model is able to map nodes with the same label to an area centered around the expectation in the embedding space.

\subsection{Augmentation-Free Graph Contrastive Learning (\afgcl)}
\label{sec:Method}
\begin{algorithm}[b]
\caption{Augmentation-Free Graph Contrastive Learning (\afgcl).}
\begin{algorithmic}
\label{alg:afgcl}
\STATE \textbf{Input:} Graph neural network $f_{\theta}$, MLP $g_{\omega}$, input adjacency matrix $\mathbf{A}$, node features $\mathbf{X}$, batch size $b$, number of hops $T$, number of positive nodes $K_{pos}$.\\
\FOR{epoch $\leftarrow$ 1, 2, $\cdots$}
  \STATE 1.Obtain the node embedding, $\mathbf{H}=f_{\theta}(\mathbf{X}, \mathbf{A})$.  \\
  \STATE 2.Obtain the hidden representation, $\mathbf{Z}=g_{\omega}(\mathbf{H})$. \\
  \STATE 3.Sample $b$ nodes for seed node set $S$. \\
  \STATE 4.Construct the node pool $P$ with the $T$-hop neighbors of each node in the node set $S$.\\
  \STATE 5.Select top-$K_{pos}$ similar nodes for every $v_i \in S$ to form the positive node set $S^{i}_{pos}$. \\
  \STATE 6.Compute the contrastive objective with Eq.~\eqref{equ:the_loss} and update parameters by applying stochastic gradient.
\ENDFOR\\
\textbf{return} Final model $f_{\theta}$.
\end{algorithmic}
\end{algorithm}
The above theoretical analysis reveals that, for each class, the embedding obtained from neighbor aggregation will concentrate toward the expectation of embedding belonging to the class. Inspired by this, we design the self-supervision signal based on the obtained embedding and propose a novel augmentation-free graph contrastive learning algorithm, \afgcl, which selects similar nodes as positive node pairs. As shown in the previous analysis, the concentration property is independent with homophily and heterophily assumptions, therefore 
\afgcl\ generalizes well on both homophilic and heterophilic graphs. And the augmentation-free design makes \afgcl\ get rid of the commonly adapted dual-branch design~\cite{zhu2021empirical} and significantly reduce the computational overhead.

As shown in Figure~\ref{fig:model}, in each iteration, the proposed framework first encodes the graph with a graph encoder $f_{\theta}$ denoted by $\mathbf{H}=f_\theta(\mathbf{X}, \mathbf{A})$. Notably, our framework allows various choices of the network architecture without any constraints. 
Then, a MLP projection head with L2-normalization, $g_{\omega}$, is employed to project the node embedding into the hidden representation $\mathbf{Z} = g_{\omega}(\mathbf{H})$. 
At each iteration, $b$ nodes are sampled to form the seed node set $S$; and their surrounding $T$-hop neighbors consist the node pool, $P$. For each seed node $v_i \in S$, the top-$K_{pos}$ nodes with highest similarity from the node pool are selected as positive set for it, and denote as $S^{i}_{pos} = \{v_i, v_{i}^1, v_{i}^2, \ldots, v_{i}^{K_{pos}} \}$. Specifically, 
\begin{equation}
\begin{aligned}
v_{i}^1, v_{i}^2, \ldots, v_{i}^{K_{pos}} = \argmax_{v_j \in P} \left( \mathbf{Z}_i^{\top} \mathbf{Z}_j, K_{pos}\right),
\end{aligned}
\end{equation}
where $\argmax(\cdot,  K_{pos})$ denotes the operator for the top-$K_{pos}$ selection, and because the hidden representation is normalized, the inner product of hidden representations is equal to the cosine similarity.
The framework is optimized with the following objective:
\begin{equation}
\label{equ:the_loss}
\begin{aligned}
\mathcal{L}_{gcl} = -2~\mathbb{E}_{v_i \sim Uni(\mathcal{V})\atop v_{i^+} \sim Uni(S^{i}_{pos})}
\left[ \mathbf{Z}_i^{\top} \mathbf{Z}_{i^+} \right]  + \mathbb{E}_{v_j \sim Uni(\mathcal{V}) \atop v_k \sim Uni(\mathcal{V})}
\left[ \left( \mathbf{Z}_j^{\top} \mathbf{Z}_{k} \right)^2 \right].
\end{aligned}
\end{equation}
where the node $v_{i^+}, v_j $ and $v_k$ are uniformly sampled from their corresponding set.
Overall, the training algorithm \afgcl\ is summarized in Algorithm~\ref{alg:afgcl}.

%% file: sec-theory.tex
\section{Theoretical Analyses}
\label{theo}
In this section, we aim to derive a performance guarantee for \afgcl. 
Note that we leave all the proof in the appendix.
First, we introduce the concept of \emph{transformed graph} as follows, which is constructed based on the original graph and the selected positive pairs. 



\begin{defi} [Transformed Graph]
\label{def:transformed_graph}
Given the original graph $\mathcal{G}$ and its node set $\mathcal{V}$, the transformed graph, $\widehat{\mathcal{G}}$, has the same node set $\mathcal{V}$ but with the selected positive pairs by \afgcl\ as the edge set, $\widehat{\mathcal{E}} = \bigcup_{i} \{ (v_i, v_{i}^k)|_{k=1}^{K_{pos}} \}$.
\end{defi}
\begin{figure}[h]
\centering
\includegraphics[width=8.0cm]{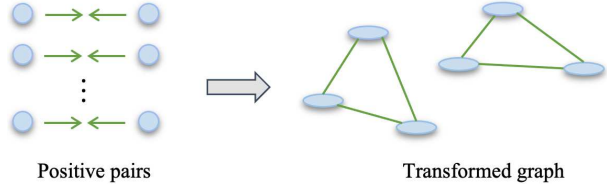}
\vspace{-1mm}
\caption{Transformed Graph formed with Positive Pairs.}
\label{fig:transformed_graph}
\vspace{-1.2mm}
\end{figure}
We also illustrate the transformed graph with Figure~\ref{fig:transformed_graph}. And we denote the adjacency matrix of transformed graph as as $\widehat{\mathbf{A}} \in\{0,1\}^{N \times N} $, the number of edges as $\hat{E}$, and the symmetric normalized matrix as $\widehat{\mathbf{A}}_{sym}=\widehat{\mathbf{D}}^{-1/2} \widehat{\mathbf{A}} \widehat{\mathbf{D}}^{-1/2}$.
Then we show that optimizing a model with the contrastive loss (Equation~\eqref{equ:the_loss}) is equivalent to the matrix factorization over the transformed graph, as stated in the following lemma:

\begin{lem} \label{lem:equiv}
Denote the learnable embedding for matrix factorization as $\mathbf{F}\in\mathbb{R}^{N\times K}$. Let $\mathbf{F}_{i} = f_{\theta}\circ g_{\omega}(v_i)$. Then, the matrix factorization loss function $\mathcal{L}_{\mathrm{mf}}$ is equivalent to the contrastive loss (Equation~\eqref{equ:the_loss}), up to an additive constant: 
\begin{equation}
\label{equ:loss_equiv}
\begin{small}
\begin{aligned}
&\mathcal{L}_{\mathrm{mf}}(\mathbf{F}) = \left\|\widehat{\mathbf{A}}_{sym}-\mathbf{F} \mathbf{F}^{\top}\right\|_{F}^{2} = \mathcal{L}_{gcl} + \text{const} 
\end{aligned}
\end{small}
\end{equation}
\end{lem}

The above lemma bridges the graph contrastive learning and the graph matrix factorization and therefore allows us to provide the performance guarantee of \afgcl\ by leveraging the power of matrix factorization. 
In addition, we provide an analysis for the inner product of hidden representations, namely $\mathbf{Z}_i^\top \mathbf{Z}_j$:

\begin{thm} \label{thm:homo_sim}
Consider a graph $G$ following the graph data assumption and Eq. (\ref{eq:feature}). Then with probability at least $1-\delta$ we have,

\begin{equation}
\big| \mathbf{Z}_i^\top \mathbf{Z}_j - \mathbb{E}[\mathbf{Z}_i^\top \mathbf{Z}_j] \big| \le \sqrt{ \frac{\sigma^2_{\max}(\mathbf{W}^\top \mathbf{W})  \log(2N^2/\delta)}{2D^2 \|\mathbf{x}^2 \|_{\psi_1}} }
\end{equation}
where $D =\min_i \mathbf{D}_{ii}  $ and sub-exponential norms $\| \mathbf{x}^2 \|_{\psi_1} = \min_i \| \mathbf{x}_{i,d}^2 \|_{\psi_1}$ for $d \in [1,F]$.
\end{thm}
By above theorem, we demonstrate that inner product of hidden representations approximates to its expectation with a high probability. Furthermore, suppose that the expected homophily over distribution of graph feature and label, i.e., $y \sim P({y_i}), \mathbf{x} \sim P_y(\mathbf{x})$, through similarity selection satisfies $\mathbb{E} [h_{edge}(\mathcal{\hat G})] = 1 - \bar \phi$. Here $\bar \phi = \mathbb{E}_{y \sim P({y_i}), \mathbf{x} \sim P_y(\mathbf{x})}[y_i \neq y_j]$. Then combining with Lemma~\ref{lem:equiv} and Theorem~\ref{thm:homo_sim}, we can now provide a theoretical guarantee for \afgcl:





\begin{thm}
\label{thm:gcl_bound}
Let $f^{*}_{\mathrm{gcl}} \in \arg \min _{f: \mathcal{X} \rightarrow \mathbb{R}^{K}}$ be a minimizer of the GCL loss, $\mathcal{L}_{\mathrm{gcl}}$.
Then there exists a linear classifier $\mathbf{B}^{*} \in \mathbb{R}^{c \times K}$ with norm $\left\|\mathbf{B}^{*}\right\|_{F} \leq 1 /\left(1-\lambda_{K}\right)$ such that, with a probability at least $1-\delta$
\begin{equation}
\begin{small}
\begin{aligned}
\mathbb{E}_{v_i}  \left[ \left\|  \vec{y}_i-\mathbf{B}^{*} f_{\mathrm{gcl}}^{*}(v)\right\|_{2}^{2} \right]
\leq \frac{\bar \phi }{\hat{\lambda}_{K+1}} + \sqrt{ \frac{\sigma^2_{\max}(\mathbf{W}^\top \mathbf{W})  \log(2N^2/\delta)}{2D^2 \|\mathbf{x}^2 \|_{\psi_1}\hat{\lambda}^2_{K+1}} },
\end{aligned}
\end{small}
\end{equation}
$\hat{\lambda}_{i}$ are the $i$ smallest eigenvalues of the symmetrically normalized Laplacian matrix of the transformed graph.
\end{thm}
\textbf{Interpreting the bound.}
The above theorem implies that if the transformed graph has a larger homophily degree (smaller ${\bar \phi}$), the bound of the prediction error will be lower. In other words, if the percentage of two nodes in positive pairs belonging to the same class is higher, the pre-trained GNN model tends to have better performance. Besides, the theorem reveals a trend that with the increase of hidden representation dimension $K$, a lower bound will be obtained.





%% file: sec-exp.tex
\section{Experiments}
\begin{table}[t]
\small
\centering
\caption{Graph Contrastive Learning on Homophilic Graphs. The highest performance of unsupervised models is highlighted in boldface. \text{OOM} indicates Out-Of-Memory on a 32GB GPU.} 
\label{tab:homo_result}
\scalebox{0.8}{
\setlength{\tabcolsep}{3.5pt}{
\begin{tabular}{l | cccccccc}
\toprule
\textbf{Model} & \textbf{Cora} & \textbf{CiteSeer} & \textbf{PubMed} & \textbf{WikiCS} & \textbf{Amz-Comp.} & \textbf{Amz-Photo} & \textbf{Coauthor-CS} & \textbf{Coauthor-Phy.} \\
\hline
MLP  & 47.92 $\pm$ 0.41 & 49.31 $\pm$ 0.26 & 69.14 $\pm$ 0.34 & 71.98 $\pm$ 0.42 & 73.81 $\pm$ 0.21 & 78.53 $\pm$ 0.32 & 90.37 $\pm$ 0.19 & 93.58 $\pm$ 0.41 \\
GCN & 81.54 $\pm$ 0.68 & 70.73 $\pm$ 0.65 & 79.16 $\pm$ 0.25 & 93.02 $\pm$ 0.11 & 86.51 $\pm$ 0.54 & 92.42 $\pm$ 0.22 & 93.03 $\pm$ 0.31 & 95.65 $\pm$ 0.16 \\
\hline
DeepWalk  & 70.72 $\pm$ 0.63 & 51.39 $\pm$ 0.41 & 73.27 $\pm$ 0.86 & 74.42 $\pm$ 0.13 & 85.68 $\pm$ 0.07 & 89.40 $\pm$ 0.11 & 84.61 $\pm$ 0.22 & 91.77 $\pm$ 0.15 \\
Node2cec  & 71.08 $\pm$ 0.91 & 47.34 $\pm$ 0.84 & 66.23 $\pm$ 0.95 & 71.76 $\pm$ 0.14 & 84.41 $\pm$ 0.14 & 89.68 $\pm$ 0.19 & 85.16 $\pm$ 0.04 & 91.23 $\pm$ 0.07 \\
\hline
GAE & 71.49 $\pm$ 0.41 & 65.83 $\pm$ 0.40 & 72.23 $\pm$ 0.71 & 73.97 $\pm$ 0.16 & 85.27 $\pm$ 0.19 & 91.62 $\pm$ 0.13 & 90.01 $\pm$ 0.71 & 94.92 $\pm$ 0.08 \\
VGAE & 77.31 $\pm$ 1.02 & 67.41 $\pm$ 0.24 & 75.85 $\pm$ 0.62 & 75.56 $\pm$ 0.28 & 86.40 $\pm$ 0.22 & 92.16 $\pm$ 0.12 & 92.13 $\pm$ 0.16 & 94.46 $\pm$ 0.13 \\
DGI  & 82.34 $\pm$ 0.71 & 71.83 $\pm$ 0.54 & 76.78 $\pm$ 0.31 & 75.37 $\pm$ 0.13 & 84.01 $\pm$ 0.52 & 91.62 $\pm$ 0.42 & 92.16 $\pm$ 0.62 & 94.52 $\pm$ 0.47 \\
GMI  & 82.39 $\pm$ 0.65 & 71.72 $\pm$ 0.15 & 79.34$ \pm$ 1.04 & 74.87 $\pm$ 0.13 & 82.18 $\pm$ 0.27 & 90.68 $\pm$ 0.18 & \text{OOM} & \text{OOM} \\
MVGRL & {\bf 83.45} $\pm$ {\bf 0.68} & {\bf 73.28} $\pm$ {\bf 0.48} & 80.09 $\pm$ 0.62 & 77.51 $\pm$ 0.06 & 87.53 $\pm$ 0.12 & 91.74 $\pm$ 0.08 & 92.11 $\pm$ 0.14 & 95.33 $\pm$ 0.05 \\
GRACE & 81.92 $\pm$ 0.89 & 71.21  $\pm$ 0.64  & {\bf 80.54} $\pm$ {\bf 0.36} & 78.19 $\pm$ 0.10 & 86.35 $\pm$ 0.44 & 92.15 $\pm$ 0.25 & {92.91 $\pm$ 0.20} & 95.26 $\pm$ 0.22 \\
GCA   & 82.07 $\pm$ 0.10 & 71.33 $\pm$ 0.37 & 80.21 $\pm$ 0.39 & 78.40 $\pm$ 0.13 & 87.85 $\pm$ 0.31 & 92.49 $\pm$ 0.11 & 92.87 $\pm$ 0.14 & {95.68 $\pm$ 0.05}  \\
BGRL & 81.44 $\pm$ 0.72 & 71.82 $\pm$ 0.48 & 80.18 $\pm$ 0.63  & 76.96 $\pm$ 0.61 & 89.62 $\pm$ 0.37 & 93.07 $\pm$ 0.34 & 92.67 $\pm$ 0.21  & 95.47 $\pm$ 0.28 \\
\midrule 
\afgcl & 83.16 $\pm$ 0.13 & 71.96 $\pm$ 0.42 & 79.05 $\pm$ 0.75 & {\bf 79.01} $\pm$ {\bf 0.51} & \textbf{{89.68} $\pm$ {0.19}} & {92.49 $\pm$ 0.31} & {91.92 $\pm$ 0.10} & 95.12 $\pm$ 0.15 \\
\bottomrule
\end{tabular}}
}
\label{tab:IR_performance}
\end{table}

\begin{table}[t]
\centering
\small
\caption{Graph Contrastive Learning on Heterophilic Graphs. The highest performance of unsupervised models is highlighted in boldface. \text{OOM} indicates Out-Of-Memory on a 32GB GPU.} 
\label{tab:heter_result}
\setlength{\tabcolsep}{5pt}{
\begin{tabular}{l | cccccc}
\toprule
\textbf{Model}       &  \textbf{Chameleon}    & \textbf{Squirrel}     & \textbf{Actor} & \textbf{Twitch-DE}    & \textbf{Twitch-gamers}         & \textbf{Genius}     \\ 
\hline
MLP          & 47.59 $\pm$ 0.73 & 31.67 $\pm$ 0.61 & 35.93 $\pm$ 0.61  & 69.44 $\pm$ 0.67 & 60.71 $\pm$ 0.18   & 86.62 $\pm$ 0.11 \\
GCN          & 66.45 $\pm$ 0.48 & 53.03 $\pm$ 0.57 & 28.79 $\pm$ 0.23  & 73.43 $\pm$ 0.71 & 62.74 $\pm$ 0.03   & 87.72 $\pm$ 0.18  \\ 
\hline
DeepWalk     & 43.99 $\pm$ 0.67 & 30.90 $\pm$ 1.09 & 25.50 $\pm$ 0.28  & 70.39 $\pm$ 0.77 & 61.71 $\pm$ 0.41 & 68.98 $\pm$ 0.15 \\ 
Node2cec     & 31.49 $\pm$ 1.17 & 27.64 $\pm$ 1.36 & 27.04 $\pm$ 0.56  & 70.70 $\pm$ 1.15 & 61.12 $\pm$ 0.29 & 67.96 $\pm$ 0.17 \\ 
\hline
GAE          & 39.13 $\pm$ 1.34 & 34.65 $\pm$ 0.81 & 25.36 $\pm$ 0.23  & 67.43 $\pm$ 1.16 & 56.26 $\pm$ 0.50 & 83.36 $\pm$ 0.21 \\ 
VGAE         & 42.65 $\pm$ 1.27 & 35.11 $\pm$ 0.92 & 28.43 $\pm$ 0.57  & 68.62 $\pm$ 1.82 & 60.70 $\pm$ 0.61 &  85.17 $\pm$ 0.52 \\ 
DGI          & 60.27 $\pm$ 0.70 & 42.22 $\pm$ 0.63 & 28.30 $\pm$ 0.76  & 72.77 $\pm$ 1.30 & 61.47 $\pm$ 0.56 &  86.96 $\pm$ 0.44 \\ 
GMI          & 52.81 $\pm$ 0.63 & 35.25 $\pm$ 1.21 & 27.28 $\pm$ 0.87  & 71.21 $\pm$ 1.27 & OOM & OOM \\ 
MVGRL        & 53.81 $\pm$ 1.09 & 38.75 $\pm$ 1.32 & 32.09 $\pm$ 1.07  & 71.86 $\pm$ 1.21 & OOM & OOM \\ 
GRACE        & 61.24 $\pm$ 0.53 & 41.09 $\pm$ 0.85 & 28.27 $\pm$ 0.43  & 72.49 $\pm$ 0.72 & OOM & OOM \\ 
GCA          & 60.94 $\pm$ 0.81 & 41.53 $\pm$ 1.09 & 28.89 $\pm$ 0.50  & 73.21 $\pm$ 0.83 & OOM & OOM \\ 
BGRL         & 64.86 $\pm$ 0.63 & 46.24 $\pm$ 0.70 & 28.80 $\pm$ 0.54  & 73.31 $\pm$ 1.11 & 60.93 $\pm$ 0.32 & 86.78 $\pm$ 0.71  \\ 
\midrule
\afgcl   & {\bf 65.28 $\pm$ 0.53}  & {\bf 52.10 $\pm$ 0.67} & {\bf 28.94 $\pm$ 0.69} & {\bf 73.51 $\pm$ 0.97} & \bf{62.04 $\pm$ 0.17} & {\bf 90.06 $\pm$ 0.18} \\ 
\bottomrule
\end{tabular}}
\end{table}

\label{sec:exp}
By extensive experiments, we show the efficacy, efficiency, and scalability of \afgcl\ for both homophilic and heterophilic graphs. 
The results on homophilic and heterophilic graph benchmarks are presented in Section~\ref{subsec:homo_result} and Section~\ref{subsec:heter_result} respectively. The scalability and complexity analysis are given in Section~\ref{subsec:memo_analysis}. In Section~\ref{subsec:dim_analysis}, we analyze the effect of the hidden dimension size.
Experiment details are given in Appendix~\ref{apd:model_detail}.

\textbf{Datasets.}
We analyze the quality of representations learned by \afgcl\ on transductive node classification benchmarks.
Specifically, we evaluate the performance of using the pretraining representations on 8 benchmark homophilic graph datasets, namely, Cora, Citeseer, Pubmed~\cite{kipf2016semi} and Wiki-CS, Amazon-Computers, Amazon-Photo, Coauthor-CS, Coauthor-Physics~\cite{shchur2018pitfalls}, as well as 6 heterophilic graph datasets, namely, Chameleon, Squirrel, Actor~\cite{pei2020geom}, Twitch-DE, Twitch-gamers, and Genius~\cite{lim2021new}.
The datasets are collected from real-world networks from different domains; their detailed statistics are summarized in Table~\ref{tab:data_stat} and the detailed descriptions are in Appendix~\ref{apd:dataset_section}.

\textbf{Baselines.}
We consider representative baseline methods belonging to the following three categories (1) Traditional unsupervised graph embedding methods, including DeepWalk~\cite{perozzi2014deepwalk} and Node2Vec~\cite{grover2016node2vec} , (2) Self-supervised learning algorithms with graph neural networks including Graph Autoencoders (GAE, VGAE)~\cite{kipf2016variational} , Deep Graph Infomax (DGI)~\cite{velickovic2019deep} , Graphical Mutual Information Maximization (GMI)~\cite{peng2020graph}, and Multi-View Graph Representation Learning (MVGRL)~\cite{hassani2020contrastive}, graph contrastive representation learning (GRACE)~\cite{zhu2020deep} Graph Contrastive learning with Adaptive augmentation (GCA)~\cite{zhu2021graph}, Bootstrapped Graph Latents (BGRL)~\cite{thakoor2021bootstrapped}, (3) Supervised learning and Semi-supervised learning, e.g., Multilayer Perceptron (MLP) and Graph Convolutional Networks (GCN)~\cite{kipf2016semi}, where they are trained in an end-to-end fashion.

\textbf{Protocol.}
We follow the evaluation protocol of previous state-of-the-art graph contrastive learning approaches.
Specifically, for every experiment, we employ the linear evaluation scheme as introduced in ~\cite{velickovic2019deep}, where each model is firstly trained in an unsupervised manner; then, the pretrained representations are used to train and test via a simple linear classifier.
For the datasets that came with standard train/valid/test splits, we evaluate the models on the public splits. For datasets without standard split, e.g., Amazon-Computers, Amazon-Photo, Coauthor-CS, Coauthor-Physics, we randomly split the datasets, where 10\%/10\%/80\% of nodes are selected for the training, validation, and test set, respectively.
For most datasets, we report the averaged test accuracy and standard deviation over 10 runs of classification. While, following the previous works~\cite{lim2021new, lim2021large}, we report the test ROC AUC on genius and Twitch-DE datasets.

\textbf{Implementation.}
We employ a two-layer GCN~\cite{kipf2016semi} as the encoder for all baselines due to its simplicity. 
Note, although the GCN will encourage the learning of low-frequency information~\cite{balcilar2020analyzing}, Ma el al.~\cite{ma2021homophily} demonstrated that GCN is enough to capture the information within heterophilic graphs following our graph assumption.
Further, the propagation for a single layer GCN is given by,
\begin{equation*}
\mathrm{GCN}_{i}(\mathbf{X}, \mathbf{A})=\sigma\left(\bar{\mathbf{D}}^{-\frac{1}{2}} \bar{\mathbf{A}} \bar{\mathbf{D}}^{-\frac{1}{2}} \mathbf{X} \mathbf{W}_{i}\right),
\end{equation*}
where $\bar{\mathbf{A}}=\mathbf{A}+\mathbf{I}$ is the adjacency matrix with self-loops, 
$\bar{\mathbf{D}}$ is the degree matrix, 
$\sigma$ is a non-linear activation function, such as ReLU, and $\mathbf{W}_{i}$ is the learnable weight matrix for the $i$'th layer. 
The proposed contrastive loss (Equation~\eqref{equ:the_loss}) is in expectation format. Its empirical version can be written as,
\begin{equation}
\label{equ:emp_loss}
\begin{aligned}
\widehat{\mathcal{L}} = -\frac{2}{N\cdot K_{pos}}\sum_{i}^N \sum_{i^+}^{K_{pos}} \left[ \mathbf{Z}_i^{\top} \mathbf{Z}_{i^+} \right]  
+ \frac{1}{N\cdot K_{neg}} \sum_{j}^N \sum_{k}^{K_{neg}}
\left[ \left( \mathbf{Z}_j^{\top} \mathbf{Z}_{k} \right)^2 \right],
\end{aligned}
\end{equation}
where to approximate the expectation over negative pairs (second term of Equation~\eqref{equ:the_loss}), we sample $K_{neg}$ nodes for each node. Notably, the empirical contrastive loss is an unbiased estimation of the Equation~\eqref{equ:the_loss}.

\subsection{Performance on Homophilic Graph}
\label{subsec:homo_result}
The homophilic graph benchmarks have been studied by several previous works~\cite{velickovic2019deep, peng2020graph, hassani2020contrastive, zhu2020deep, thakoor2021bootstrapped}.
We re-use their configuration and compare \afgcl\ with those methods.
The result is summarized in Table~\ref{tab:homo_result}.
The augmentation-based GCL methods can outperform the corresponding supervised training model. 
As we analyzed in Section~\ref{sec:aug_study}, those methods implicitly perturb the high-frequency information across different augmentation views, and the commonly adopted InfoNCE loss~\cite{van2018representation} enforce the target GNN model to capture the low-frequency information by enforcing the learned representation invariant with different augmentations. And the low-frequency information contributes to the success of the previous GCL methods on the homophilic graphs, which is aligned with previous analysis~\cite{nt2019revisiting, balcilar2020analyzing}.
Compare with augmentation-based GCL methods, \afgcl\ outperforms previous methods on two datasets and achieves competitive performance on the other datasets, which shows the effectiveness of the augmentation-free design on homophilic graphs. Note that in our analysis, these baselines are indeed tailored for homophilic graphs and \afgcl\ is a theoretically-justified contrastive learning framework without augmentations.

\subsection{Performance on Heterophilic Graph}
\label{subsec:heter_result}
We further assess the model performance on heterophilic graph benchmarks that introduced by Pei et al.~\cite{pei2020geom} and Lim et al.~\cite{lim2021large}. 
As shown in Table~\ref{tab:heter_result}, \afgcl\ achieves the best performance on 6 of 6 heterophilic graphs by an evident margin. We notice that for the two graphs with the lowest homophily degree, Chameleon and Squirrel, \afgcl\ outperform the previous methods with a large margin. The result verifies that our proposed method is suitable for heterophilic graphs. Interestingly, some of the baselines even cannot scale to some graphs and perform poorly on the others. We believe it is due to the high computational cost and loss of the high-frequency information after graph augmentations, which is an innate deficiency of these methods.
\vspace{-5pt}
\begin{table}[h]
\centering
\caption{Computational requirements on a set of standard benchmark graphs. OOM indicates runing out of memory on a 32GB GPU.} 
\label{tab:memo}

\setlength{\tabcolsep}{4pt}{
\begin{tabular}{l | ccccc}
\toprule
\text {Dataset}  & \text {Coauthor-CS} & \text {Coauthor-Phy.} & \text {Genius}  & \text {Twitch-gamers} \\
\hline 
\text {\# Nodes}  & 18,333 & 34,493 & 421,961 & 168,114 \\
\text {\# Edges}  & 327,576 & 991,848 & 984,979 & 6,797,557 \\
\hline 
\text {GRACE} & 13.21 {GB} & 30.11 {GB} & {OOM}  & {OOM} \\
\text {BGRL} & 3.10 {GB} & 5.42 {GB} & {8.18} {GB} & {26.22} {GB} \\
\text {\afgcl} & 2.07 {GB} & 3.21 {GB} & 6.24 {GB} & {22.15} {GB} \\
\bottomrule
\end{tabular}}
\end{table}
\vspace{-10pt}
\subsection{Computational Complexity Analysis}
\label{subsec:memo_analysis}
In order to illustrate the advantages of \afgcl, 
we provide a brief comparison of the time and space complexities between \afgcl, the previous strong contrastive method, GCA~\cite{zhu2021graph}, and the memory-efficient contrastive method, BGRL~\cite{thakoor2021bootstrapped}.
GCA is the advanced version of GRACE~\cite{zhu2021graph} and performs a quadratic all-pairs contrastive computation at each update step.
BGRL, inspired by the bootstrapped design of contrastive learning methods in CV~\cite{grill2020bootstrap}, conducts the pairwise comparison between the embeddings learned by an online encoder and a target encoder. Although BGRL 
does not require negative examples, the two branch design, two different encoders and four embedding table still need to be kept during training, making it hard to scale.

Consider a graph with N nodes and E edges, and a graph neural network (GNN), $f$, that compute node embeddings in time and space $O(N + E)$. This property is satisfied by most popular GNN architectures, e.g., convolutional~\cite{kipf2016semi}, attentional~\cite{velivckovic2017graph}, or message-passing~\cite{gilmer2017neural} networks and have been analyzed in the previous works~\cite{thakoor2021bootstrapped}.
BGRL performs four GNN computations per update step, in which twice for the target and online encoders, and twice for each augmentation, and a node-level projection; GCA performs two GNN computations (once for each augmentation), plus a node-level projection. Both methods backpropagate the learning signal twice (once for each augmentation), and we assume the backward pass to be approximately as costly as a forward pass. 
Both of them will compute the augmented graphs by feature masking and edge masking on the fly, the cost for augmentation computation is nearly the same. Thus the total time and space complexity per update step for BGRL is
$6C_{encoder}(E + N) + 4C_{proj}N + C_{prod}N + C_{aug}$ and $4 C_{encoder}(E + N) + 4C_{proj}N + C_{prod}N^2 + C_{aug}$ for GCA. The $C_{prod}$ depends on the dimension of node embedding and we assume the node embeddings of all the methods with the same size.
For our proposed method, only one GNN encoder is employed and we compute the inner product of $b$ nodes to construct positive samples and $K_{pos}$ and $K_{neg}$ inner product for the loss computation. Then for \afgcl, we have:
$2C_{encoder}(E + N) + 2C_{proj}N + C_{prod}(K_{pos}+K_{neg})^2$. 
We empirically measure the peak of GPU memory usage of \afgcl, GCA and BGRL. As a fair comparison, we set the embedding size as 128 for all those methods on the four datasets and keep the other hyper-parameters of the three methods the same as the main experiments.
The result is summarized in Table~\ref{tab:memo}.
\vspace{-5pt}
\begin{table}[h]
\centering
\caption{The performance of \afgcl\ with different hidden dimension. The average accuracy over 10 runs is reported.} 
\label{tab:dim_study}
\begin{tabular}{l|cccc}
\toprule
       & WikiCS & Amz-Comp. & Actor & Twitch-DE \\
\hline
$K=256$  &    77.11    &    87.79       & 27.17      & 71.50  \\
$K=512$  &   78.14     &    88.21       & 27.40      & 72.39  \\
$K=1024$ &   79.01     &    89.68       & 28.94      &  73.51 \\
\bottomrule
\end{tabular}
\end{table}
\subsection{Representation Size Analysis}
\label{subsec:dim_analysis}
As implied by Theorem~\ref{thm:gcl_bound}, a larger hidden dimension leads to better performance. We empirically verify that on four datasets. The result is summarized in Table~\ref{tab:dim_study} and we can see that the performance increases consistently with larger $K$.

%% file: sec-conclusion.tex
\section{Conclusion}
\label{sec:conclusion}
In this work, we first investigate the effect of graph augmentation techniques--a crucial part of existing graph contrastive learning algorithms. Specifically, they tend to preserve the low-frequency information and perturb the high-frequency information, which mainly contributes to the success of augmentation-based GCL algorithms on the homophilic graph, but limits its application on the heterophilic graphs. Then, motivated by our theoretical analyses of the features aggregated by Graph Neural Networks, we propose an augmentation-free graph contrastive learning method, \afgcl, wherein the self-supervision signal is constructed based on the aggregated features. 
We further provide the theoretical guarantee for the performance of \afgcl\ as well as the analysis of its efficacy.
Empirically, we show that \afgcl\ can outperform or be competitive with SOTA methods on 8 homophilic graph benchmarks and 6 heterophilic graph benchmarks with significantly less computational overhead.
Admittedly, we mainly focus on the node classification problem. We would like to leave the exploration of regression problem and graph classification problem in the future.

%% file: sec-appendix.tex
\appendix
\section{Dataset Information}
\label{apd:dataset_section}
We evaluate our models on eight node classification homophilic benchmarks: Cora, Citeseer, Pubmed, WikiCS, Coauthor-CS, Coauthor-Physics, Amazon-Computer and Amazon-Photo and seven non-homophilic benchmarks: Chameleon, Squirrel, Deezer, Penn94 (FB100), Twitch-gamers, Twitch-DE, Genius. 
The datasets are collected from real-world networks from different domains and we provide dataset statistics in Table~\ref{tab:data_stat}.  For the 8 homophilic graph data, we use the processed version provided by PyTorch Geometric~\cite{fey2019fast}. Besides, for the 6 heterophilic graph data, 3 of them, e.g., Chameleon, Squirrel and Actor are provided by PyTorch Geometric. The other three dataset, genius, twitch-DE and twitch-gamers can be obtained from the official github repository\footnote{\url{https://github.com/CUAI/Non-Homophily-Large-Scale}}, in which the standard splits for all the 6 heterophilic graph datasets can also be obtained.
\begin{table}[h]
\centering
\small
\caption{Statistics of datasets used in experiments.} 
\label{tab:data_stat}
\setlength{\tabcolsep}{2.5pt}{
\begin{tabular}{l | rrrrrr}
\toprule
\text { Name } & \text { Nodes } & \text { Edges } & \text { Classes } & \text { Feat. }  & $h_{node}$ &  $h_{edge}$ \\
\hline
\text { Cora } & 2,708 & 5,429 & 7 & 1,433 & .825 & .810 \\
\text { CiteSeer } & 3,327 & 4,732 & 6 & 3,703 & .717 & .736 \\
\text { PubMed } & 19,717 & 44,338 & 3 & 500 & .792 & .802 \\
\text { Coauthor CS } & 18,333 & 327,576 & 15 & 6,805 & .832 & .808 \\
\text { Coauthor Phy. } & 34,493 & 991,848 & 5 & 8,451 & .915 & .931 \\
\text { Amazon Comp. } & 13,752 & 574,418 & 10 & 767  & .785 & .777 \\
\text { Amazon Photo } & 7,650 & 287,326 & 8 & 745  & .836 & .827  \\
\text { WikiCS } & 11,701 & 216,123 & 10 & 300 & .658 & .654 \\
\hline  
\text { Chameleon } & 2,277 & 36,101 & 5 & 2,325  & .103 & .234 \\
\text { Squirrel } & 5,201 & 216,933 & 5 & 2,089 & .088 & .223 \\
\text { Actor } & 7,600 &  33,544 & 5 & 9,31 & .154 & .216 \\
\text { Twitch-DE } & 9,498 & 153,138 & 2 & 2,514  & .529 & .632 \\
\text { Twitch-gamers } & 168,114 & {\bf 6,797,557} & 2 & 7 & .552 & .545 \\
\text { Genius } & {\bf 421,961} & 984,979 & 2 & 12 & .477 & .618 \\
\bottomrule
\end{tabular}}
\end{table}

\vspace{-5pt}
\section{Implementation Details}
\label{apd:model_detail}
\textbf{Model Architecture and hyperparamters.}
As we described in Section~\ref{sec:exp}, we employ a two-layer GCN~\cite{kipf2016semi} as the encoder for all methods. Following the previous works~\cite{kipf2016semi, lim2021new, lim2021large}, we apply the l2-normalization for the raw features before training and use batch normalization within the graph encoder.
The hyperparameters setting for all experiments are summarized in Table~\ref{tab:hyper_args}.
We would like to release our code after acceptance. \\
\textbf{Linear evaluation of embeddings.}
In the linear evaluation protocol, the final evaluation is over representations obtained from pretrained model.
When fitting the linear classifier on top of the frozen learned embeddings, the gradient will not flow back to the encoder. We optimize the one layer linear classifier 1000 epochs using Adam with learning rate 0.0005.\\
\textbf{Hardware and software infrastructures.}
Our model are implemented with PyTorch Geometric 2.0.3~\cite{fey2019fast}, PyTorch 1.9.0~\cite{paszke2017automatic}. 
We conduct experiments on a computer server with four NVIDIA Tesla V100 SXM2 GPUs (with 32GB
memory each) and twelve Intel Xeon Gold 6240R 2.40GHz CPUs.

\begin{table}[h]
\centering
\caption{Hyperparameter settings for all experiments.} 
\label{tab:hyper_args}
\small
\begin{tabular}{l|llllll}
\toprule
              & \textit{lr.} & $K_{pos}$ & $K_{neg}$ & $b$ & $T$ & $K$   \\
\hline
Cora          & .0010 & 5         & 100       & 512 & 2 & 1024 \\
CiteSeer      & .0020 & 10        & 200       & 256 & 2 & 1024 \\
PubMed        & .0010  & 5        & 100       & 256 & 2 & 1024 \\
WikiCS        & .0010 & 5         & 100       & 512 & 2 & 1024 \\
Amz-Comp.     & .0010 & 5         & 100       & 256 & 2 & 1024 \\
Amz-Photo     & .0010 & 5         & 100       & 512 & 2 & 1024 \\
Coauthor-CS   & .0010 & 5         & 100       & 512 & 2 & 1024 \\
Coauthor-Phy. & .0010 & 5         & 100       & 256 & 2 & 1024 \\
Chameleon     & .0010 & 5         & 100       & 512 & 3 & 1024 \\
Squirrel      & .0010 & 5         & 100       & 512 & 3 & 1024 \\
Actor         & .0010 & 10        & 100       & 512 & 4 & 1024 \\
Twitch-DE     & .0010 & 5        & 100       & 512  & 4 & 1024 \\
Twitch-gamers & .0010 & 5        & 100       & 256 & 4 & 128 \\
Genius        & .0010 & 5        & 100       & 256 & 3 & 512 \\
\bottomrule
\end{tabular}
\end{table}

\section{More Results for the Study of Graph Augmentation}
\label{apd:aug_study}
\begin{figure}[h]
     \centering
     \begin{subfigure}[b]{6cm}
         \centering
         \includegraphics[width=6cm]{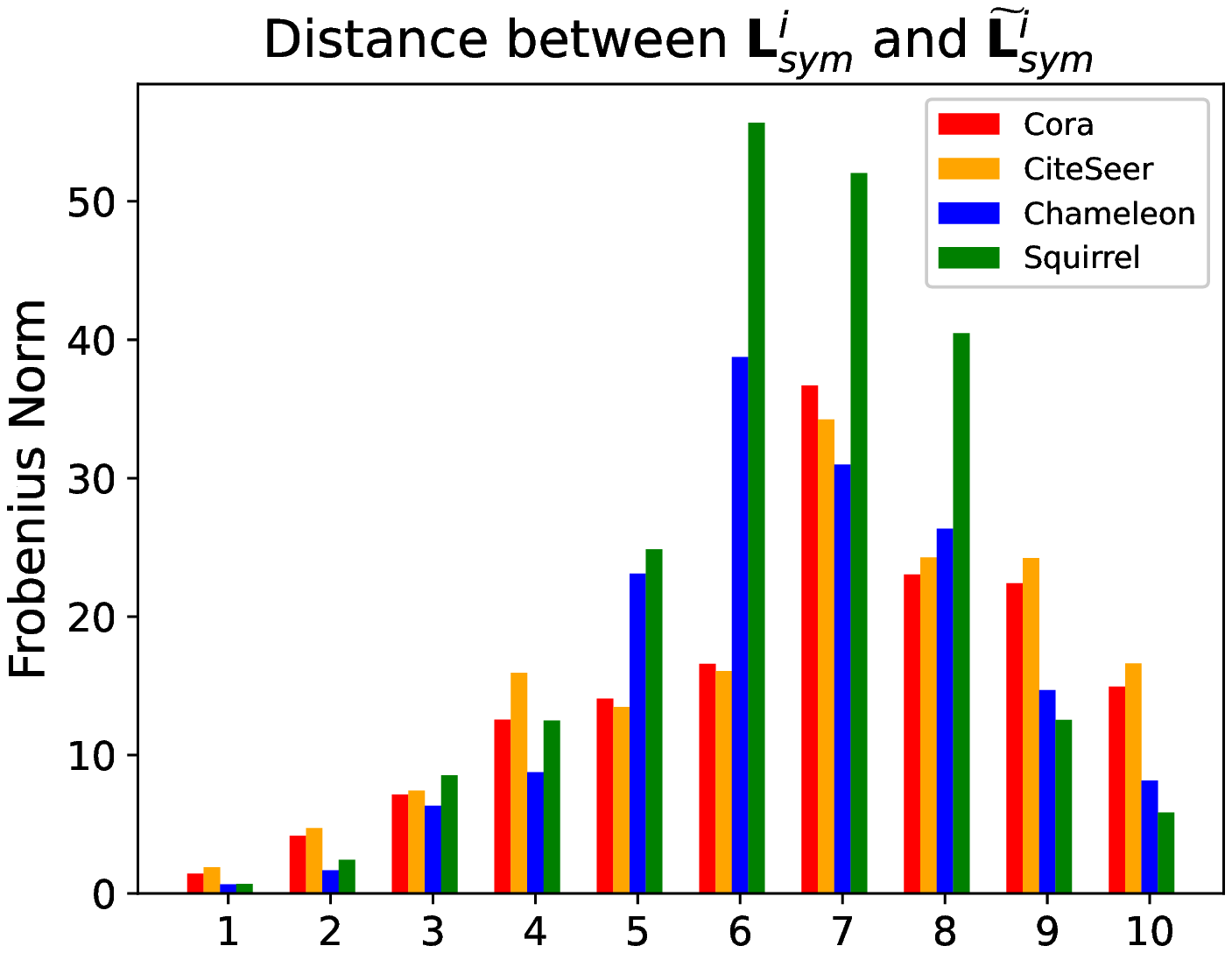}
         \caption{The effect of Edge Adding.}
         \label{fig:ea_fm_observation}
     \end{subfigure}
     \begin{subfigure}[b]{6cm}
         \centering
         \includegraphics[width=6cm]{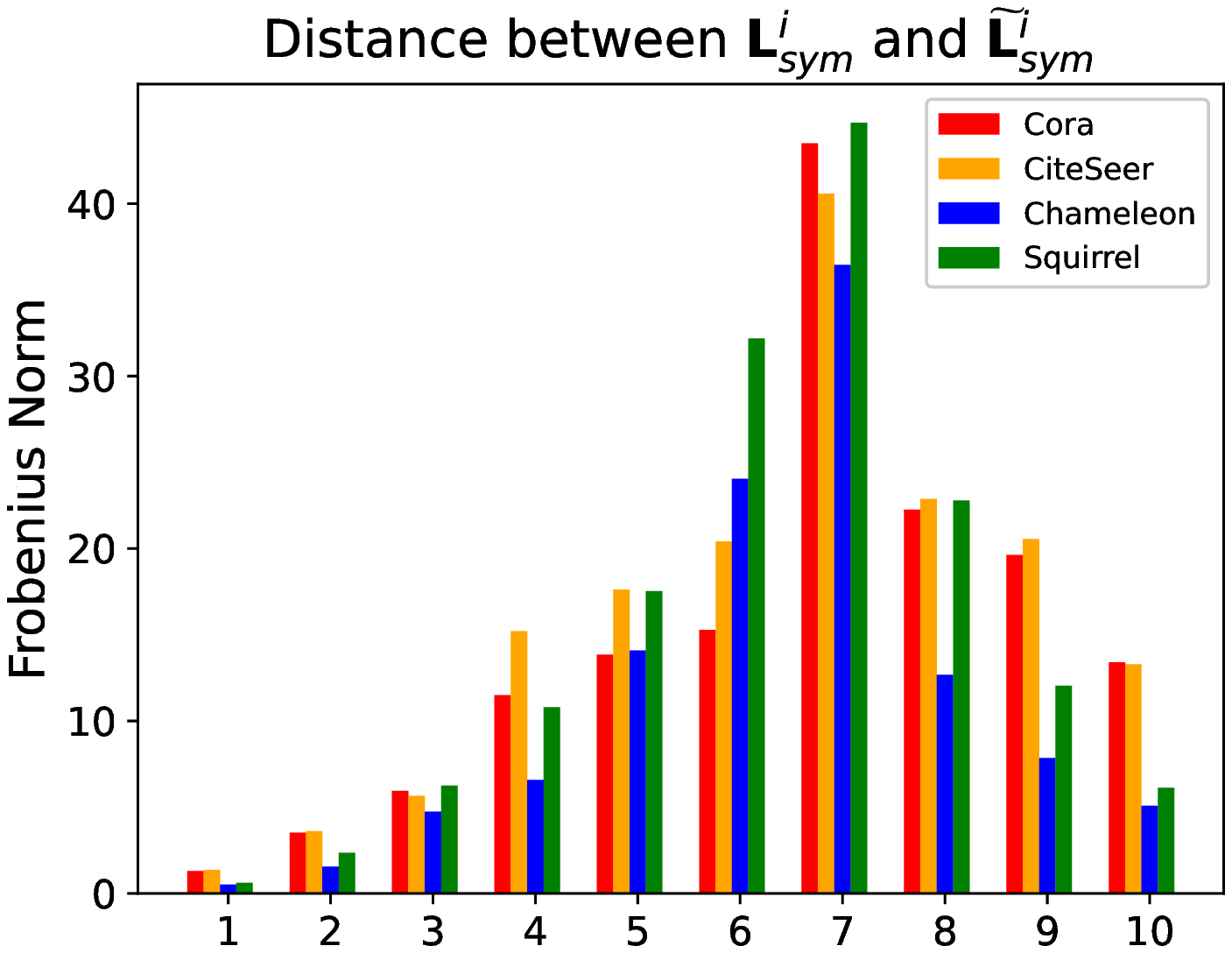}
         \caption{The effect of  Graph Diffusion.}
         \label{fig:fusion_observation}
     \end{subfigure}
\end{figure}

We decomposed the Laplacian matrix into 10 parts and compute the average Frobenius distance for each part over 10 independent runs. As shown in Figure~\ref{fig:ea_fm_observation} and Figure~\ref{fig:fusion_observation}, both the edge adding with 20\% edges and diffusion with $\alpha=0.2$ have less impact on the low frequency components.

\vspace{-5pt}
\section{Detailed Proofs}
\subsection{Proof of Lemma \ref{lem:emb}}
\label{apd: lemma_emb_proof}
{\it Proof.} 
We first calculate the expectation of aggregated embedding:
\begin{equation}
\begin{small}
\begin{aligned}
\mathbb{E}[ f_\theta(\mathbf{x}_i)] =\mathbb{E} \bigg[ \mathbf{W} \sum_{j \in \mathcal{N}(i)} \frac{1}{\mathbf{D}_{ii}  }\mathbf{x}_j \bigg] = \mathbf{W} \mathbb{E}_{y \sim P_{y_i}, \mathbf{x} \sim P_y(\mathbf{x}) } [\mathbf{x}]
\end{aligned}
\end{small}
\end{equation}
This equation is based on the graph data assumption such that $\mathbf{x}_{j} \sim P_{y_i}(\mathbf{x})$ for every $j$. Now we provide a concentration analysis. Because each feature $\mathbf{x}_i$ is a sub-Gaussian variable, then by Hoeffding's inequality, with probability at least $1-\delta'$ for each $d \in [1,F]$, we have,
\begin{equation}
\begin{small}
\begin{aligned}
\bigg | \frac{1}{\mathbf{D}_{ii}} \sum_j (\mathbf{x}_{j,d} - \mathbb{E}[\mathbf{x}_{j,d}]) \bigg | \le \sqrt{ \frac{  \log(2/\delta')}{2\mathbf{D}_{ii}\|\mathbf{x}_{j,d} \|_{\psi_2}} }
\end{aligned}
\end{small}
\end{equation}
where $\| \mathbf{x}_{j,d}\|_{\psi_2}$ is sub-Gaussian norm of $\mathbf{x}_{j,d}$. Furthermore, because each dimension of $\mathbf{x}_j$ is i.i.d., thus we define $\| \mathbf{x}_j \|_{\psi_2} = \| \mathbf{x}_{j,d} \|_{\psi_2}$ Then we apply a union bound by setting $\delta' = F \delta$ on the feature dimension $k$. Then with
probability at least $1-\delta$ we have
\begin{equation}
\begin{small}
\begin{aligned}
\bigg | \frac{1}{\mathbf{D}_{ii}} \sum_j (\mathbf{x}_{j,d} - \mathbb{E}[\mathbf{x}_{j,d}]) \bigg | \le \sqrt{ \frac{  \log(2F/\delta)}{2\mathbf{D}_{ii}\|\mathbf{x} \|_{\psi_2}} }
\end{aligned}
\end{small}
\end{equation}
Next, we use the matrix perturbation theory,
\begin{equation}
\begin{small}
\begin{aligned}
\bigg \| \frac{1}{\mathbf{D}_{ii}} \sum_j (\mathbf{x}_{j,d} - \mathbb{E}[\mathbf{x}_{j,d}]) \bigg \|_2 & \le \sqrt{F}  \bigg | \frac{1}{\mathbf{D}_{ii}} \sum_j (\mathbf{x}_{j,d} - \mathbb{E}[\mathbf{x}_{j,d}]) \bigg | \\
& \le \sqrt{ \frac{F  \log(2F/\delta)}{2\mathbf{D}_{ii}\|\mathbf{x} \|_{\psi_2}} }
\end{aligned}
\end{small}
\end{equation}
Finally, plug the weight matrix into the inequality,
\begin{equation}
\begin{small}
\begin{aligned}
\| f_\theta(\mathbf{x}_i) - \mathbb{E}[f_\theta(\mathbf{x}_i)]   \| \le \sigma_{\max}(\mathbf{W})\bigg \| \frac{1}{\mathbf{D}_{ii}} \sum_j (\mathbf{x}_{j,k} - \mathbb{E}[\mathbf{x}_{j,k}]) \bigg \|_2
\end{aligned}
\end{small}
\end{equation}
where $\sigma_{\max}$ is the largest singular value of weight matrix.

\textbf{Connection with existing observations}
In Lemma 1, we theoretically analyze the concentration property of aggregated features of graphs following the Graph Assumption. 
Note, the concentration property has also been empirically observed in some other recent works~\cite{wang2022fastgcl, trivedi2022augmentations}. 
Specifically, in the Figure 1 of Wang et al.~\cite{wang2022fastgcl}, a t-SNE visualization of node representations of the Amazon-Photo dataset shows that the representations obtained by a randomly initialized untrained SGC~\cite{wu2019simplifying} model will concentrate to a certain area if they are from the same class. This observation also provides an explanation for the salient performance achieved by AFGCL on the Amazon-Photo, as shown in Table 1. Besides, in the Figure 2 of Trivedi et al.~\cite{trivedi2022augmentations}, the concentration property has also been observed.

\subsection{Proof of Theorem \ref{thm:homo_sim}}
\label{apd: theorem_homo_sim_proof}
{\it Proof.} 
The concentration analysis is based on the result obtained in Lemma \ref{lem:emb}. We first write down the detailed expression for each pair of $i,j$,
\begin{equation}
\begin{small}
\begin{aligned}
s_{i,j} \equiv \mathbf{x}^\top_i \mathbf{W}^\top \mathbf{W} \mathbf{x}_j
\end{aligned}
\end{small}
\end{equation}
We first bound $\mathbf{x}^\top_i \mathbf{x}_j$. Because $\mathbf{x}_i$ and $\mathbf{x}_j$ are independently sampled from an identical distribution, then the product $\mathbf{x}^\top_i \mathbf{x}_j$ is sub-exponential. This can been seen from Orilicz norms relation that $\| x^2 \|_{\psi_1} = (\| x^2 \|_{\psi_2})^2$, where $\| \mathbf{x}\|_{\psi_2}$ is sub-exponential norm of $\mathbf{x}^2$. Then by the Hoeffding's inequality for sub-exponential variable, with a probability at least $1-\delta$, we have
\begin{equation}
\begin{small}
\begin{aligned}
| \mathbf{x}^\top_i \mathbf{x}_j - \mathbb{E}_{\mathbf{x}_i \sim P_{y_i} ,\mathbf{x}_j \sim P_{y_j} } [\mathbf{x}^\top_i \mathbf{x}_j] | \le \sqrt{ \frac{ \sigma^2_{\max}(\mathbf{W}^\top \mathbf{W})  \log(2/\delta)}{2\|\mathbf{x}^2 \|_{\psi_1}} }
\end{aligned}
\end{small}
\end{equation}
Because that the aggregated feature is normalized by the degree of corresponding node, we have, for each pair of $i,j$
\begin{equation}
\begin{small}
\begin{aligned}
|s_{i,j} - \mathbb{E}[s_{i,j}]| \le \sqrt{ \frac{  \log(2/\delta)\sigma^2_{\max}(\mathbf{W}^\top \mathbf{W}) }{2\|\mathbf{x}^2 \|_{\psi_1}\mathbf{D}_{ii} \mathbf{D}_{jj}}  } 
\le \sqrt{ \frac{ \sigma^2_{\max}(\mathbf{W}^\top \mathbf{W}) \log(2/\delta)}{2\|\mathbf{x}^2 \|_{\psi_1}D^2}  } 
\end{aligned}
\end{small}
\end{equation}
where $D =\min_i \mathbf{D}_{ii}$ for $i \in [1, N]$.
Finally we apply a union bound  over a pair of $i,j$. Then with
probability at least $1-\delta$ we have
\begin{equation}
\begin{small}
\begin{aligned}
\big| \mathbf{Z}_i^\top \mathbf{Z}_j - \mathbb{E}[\mathbf{Z}_i^\top \mathbf{Z}_j] \big| \le \sqrt{ \frac{\sigma^2_{\max}(\mathbf{W}^\top \mathbf{W})  \log(2N^2/\delta)}{2D^2 \|\mathbf{x}^2 \|_{\psi_1}} }
\end{aligned}
\end{small}
\end{equation}

\vspace{-10pt}
\subsection{Proof of Lemma \ref{lem:equiv}}
\label{apd:lemma_equivalence_proof}
To prove this lemma, we first introduce the concept of the probability adjacency matrix.
For the transformed graph $\widehat{\mathcal{G}}$, we denote its probability adjacency matrix as $\widehat{\mathbf{W}}$, in which $\hat{w}_{ij} = \frac{1}{\widehat{E}}\cdot \widehat{\mathbf{A}_{ij}}$. $\hat{w}_{ij}$ can be understood as the probability that two nodes have an edge and the weights sum to $1$ because the total probability mass is 1: $\sum_{i,j}\hat{w}_{i,j^{\prime}}=1$, for $v_i, v_j \in \mathcal{V}$.
Then the corresponding symmetric normalized matrix is $\widehat{\mathbf{W}}_{sym}=\widehat{\mathbf{D}}^{-1/2}_{\mathbf{w}} \widehat{\mathbf{W}} \widehat{\mathbf{D}}^{-1/2}_{\mathbf{w}}$, and the $\widehat{\mathbf{D}} = diag\big([ \hat{w}_{1}, \ldots,  \hat{w}_{N}]\big)$, where $ \hat{w}_{i} = \sum_{j}\hat{w}_{ij}$. 
We then introduce the Matrix Factorization Loss which is defined as:
\begin{equation}
\begin{small}
\label{equ:mf_loss}
\begin{aligned}
\min _{\mathbf{F} \in \mathbb{R}^{N \times k}} \mathcal{L}_{\mathrm{mf}}(\mathbf{F}):=\left\|\widehat{\mathbf{A}}_{sym}-\mathbf{F} \mathbf{F}^{\top}\right\|_{F}^{2}.
\end{aligned}
\end{small}
\end{equation}
By the classical theory on low-rank approximation, Eckart-Young-Mirsky theorem~\cite{eckart1936approximation}, any minimizer $\widehat{\mathbf{F}}$ of $\mathcal{L}_{\mathrm{mf}}(\mathbf{F})$ contains scaling of the smallest eigenvectors of $\mathbf{L}_{sym}$ (also, the largest eigenvectors of $\widehat{\mathbf{A}}_{sym}$) up to a right transformation for some orthonormal matrix $\mathbf{R} \in \mathbb{R}^{k \times k}$. We have $\widehat{\mathbf{F}}=\mathbf{F}^{*}$. $\operatorname{diag}\left(\big[\sqrt{1-\lambda_{1}}, \ldots, \sqrt{1-\lambda_{k}}\big]\right) \mathbf{R}$, where $\mathbf{F}^{*} =\left[\mathbf{u}_{1}, \mathbf{u}_{2}, \cdots, \mathbf{u}_{k}\right] \in \mathbb{R}^{N \times k}$.
To proof the Lemma~\ref{lem:equiv}, we first present the Lemma~\ref{lem:equiv_matrix_same}.
\begin{lem} \label{lem:equiv_matrix_same}
For transformed graph, its probability adjacency matrix, and adjacency matrix are equal after symmetric normalization, $\widehat{\mathbf{W}}_{sym} = \widehat{\mathbf{A}}_{sym}$.
\end{lem}
\vspace{-10pt}
{\it Proof.} 
For any two nodes $v_i, v_j \in \mathcal{V}$ and $i\neq j$, we denote the the element in $i$-th row and $j$-th column of matrix $\widehat{\mathbf{W}}_{sym}$ as $\widehat{\mathbf{W}}_{sym}^{ij}$.
\begin{equation}
\begin{small}
\begin{aligned}
{\widehat{\mathbf{W}}_{sym}}^{ij} = \frac{1}{\sqrt{\sum_{k} \hat{w}_{ik}} \sqrt{\sum_{k} \hat{w}_{kj}} } \frac{1}{E} {\widehat{\mathbf{A}}}^{ij}  = \frac{1}{\sqrt{\sum_{k} \widehat{\mathbf{A}}_{ik}} \sqrt{\sum_{k} \widehat{\mathbf{A}}_{kj} } } {\widehat{\mathbf{A}}}^{ij} = \widehat{\mathbf{A}}_{sym}^{ij}.
\end{aligned}
\end{small}
\end{equation}
By leveraging the Lemma~\ref{lem:equiv_matrix_same}, we present the proof of Lemma~\ref{lem:equiv}.

{\it Proof.} 
We start from the matrix factorization loss over $\widehat{\mathbf{A}}_{sym}$ to show the equivalence. 
\begin{equation}
\label{equ:equivalence_1}
\begin{small}
\begin{aligned}
& \quad \quad  \| \widehat{\mathbf{A}}_{sym} -\mathbf{F}\mathbf{F}^{\top} \|^2_F  = \| \widehat{\mathbf{W}}_{sym} -\mathbf{F}\mathbf{F}^{\top} \|^2_F \\
& = \sum_{ij} \big(  \frac{\hat{w}_{ij}}{\sqrt{\hat{w}_i} \sqrt{\hat{w}_j}} - {f}_{\mathrm{mf}}(v_i)^{\top} {f}_{\mathrm{mf}}(v_j) \big)^2\\
&=\sum_{ij} ({f}_{\mathrm{mf}}(v_i)^{\top}{f}_{\mathrm{mf}}(v_j))^2 -2 \sum_{ij} \frac{\hat{w}_{ij}}{\sqrt{\hat{w}_i}\sqrt{\hat{w}_j}} {f}_{\mathrm{mf}}(v_i)^{\top}{f}_{\mathrm{mf}}(v_j) + \| \hat{\mathbf{W}}_{sym} \|^2_{F} \\
\end{aligned}
\end{small}
\end{equation}
\begin{equation}
\begin{small}
\begin{aligned}
&= \sum_{ij} \hat{w}_i \hat{w}_j \big[ \big(\frac{1}{\sqrt{\hat{w}_i}} \cdot{f}_{\mathrm{mf}}(v_i)  \big)^{\top} \big( \frac{1}{\sqrt{\hat{w}_j}} \cdot{f}_{\mathrm{mf}}(v_j) \big) \big]^2 \\
&\quad\quad\quad\quad\quad-2\sum_{ij} \hat{w}_{ij} \big(\frac{1}{\sqrt{\hat{w}_i}} \cdot{f}_{\mathrm{mf}}(v_i)  \big)^{\top} \big( \frac{1}{\sqrt{\hat{w}_j}} \cdot{f}_{\mathrm{mf}}(v_j) \big) + C\\
\end{aligned}
\end{small}
\end{equation}
where ${f}_{\mathrm{mf}}(v_i)$ is the $i$-th row of the embedding matrix $\mathbf{F}$.
The $\hat{w}_i$ which can be understood as the node selection probability which is proportional to the node degree. Then, we can define the corresponding sampling distribution as $P_{deg}$.
If and only if $\sqrt{w_{i}} \cdot f_{\theta}\circ g_{\omega}(v_i)= {f}_{\mathrm{mf}}(v_i)=\mathbf{F}_i$,
we have:
\begin{equation}
\label{equ:equivalence_2}
\begin{small}
\begin{aligned}
&\mathbb{E}_{v_i \sim P_{deg} \atop v_j \sim P_{deg}}\left( f_{\theta}\circ g_{\omega} (v_i)^{\top} f_{\theta}\circ g_{\omega}(v_j) \right)^2 \\
&- 2~\mathbb{E}_{v_i \sim Uni(\mathcal{V}) \atop v_{i^+}\sim Uni(\mathcal{N}(v_i))} \left(   f_{\theta}\circ g_{\omega}(v_i)^{\top} f_{\theta}\circ g_{\omega}(v_{i^{+}}) \right) + C
\end{aligned}
\end{small}
\end{equation}
where $\mathcal{N}(v_i)$ denotes the neighbor set of node $v_i$ and $Uni(\cdot)$ is the uniform distribution over the given set. Because we constructed the transformed graph by selecting the top-$K_{pos}$ nodes for each node, then all nodes have the same degree. We can further simplify the objective as:
\begin{equation}
\label{equ:equivalence_3}
\begin{small}
\begin{aligned}
\mathbb{E}_{v_i \sim Uni(\mathcal{V}) \atop v_j \sim Uni(\mathcal{V})}\left( \mathbf{Z}_i^{\top}\mathbf{Z}_j \right)^2 - 2~\mathbb{E}_{v_i \sim Uni(\mathcal{V}) \atop v_{i^+}\sim Uni(S^i_{pos})} \left(   \mathbf{Z}_i^{\top} \mathbf{Z}_{i^{+}} \right) + C.
\end{aligned}
\end{small}
\end{equation}
Due to the node selection procedure, the factor $\sqrt{w_{i}}$ is a constant and can be absorbed by the neural network, $f_{\theta}\circ g_{\omega}$. Besides, because $\mathbf{Z}_i = f_{\theta}\circ g_{\omega}(v_i)$, we can have the Equation~\ref{equ:equivalence_3}. Therefore, the minimizer of matrix factorization loss is equivalent with the minimizer of the contrastive loss.


\subsection{Proof of Theorem \ref{thm:gcl_bound}}
\label{apd:theorem_gcl_bound_proof}

Recently, Haochen et al.~\cite{haochen2021provable} presented the following theoretical guarantee for the model learned with the matrix factorization loss.
\begin{lem}
\label{thm:orig_mf_bound}
For a graph $\mathcal{G}$, let $f^{*}_{\mathrm{mf}} \in \arg \min _{f_{\mathrm{mf}}: \mathcal{V} \rightarrow \mathbb{R}^{K}}$ be a minimizer of the matrix factorization loss, $\mathcal{L}_{\mathrm{mf}}(\mathbf{F})$, where $\mathbf{F}_i = f_{\mathrm{mf}}(v_i)$.
Then, for any label $\mathbf{y}$,
there exists a linear classifier $\mathbf{B}^{*} \in \mathbb{R}^{c \times K}$ with norm $\left\|\mathbf{B}^{*}\right\|_{F} \leq 1 /\left(1-\lambda_{K}\right)$ such that 
\begin{equation}
\begin{small}
\begin{aligned}
\mathbb{E}_{v_i}  \left[ \left\|  \vec{y}_i-\mathbf{B}^{*} f_{\mathrm{mf}}^{*}(v_i)\right\|_{2}^{2} \right]
\leq \frac{\phi^{\mathbf{y}}}{\lambda_{K+1}},
\end{aligned}
\end{small}
\end{equation}
where $\vec{y}_i$ is the one-hot embedding of the label of node $v_i$.
The difference between labels of connected data points is measured by  $\phi^{{\mathbf{y}}}$,
$
\phi^{{\mathbf{y}}}:=\frac{1}{E} \sum_{v_i, v_j \in \mathcal{V}} {\mathbf{A}}_{ij} \cdot \mathbbm{1}\left[{y}_i \neq {y}_j\right].
$ 
\end{lem}

{\it Proof.} 
This proof is a direct summary on the established lemmas in previous section. By Lemma \ref{lem:equiv} and Lemma \ref{thm:orig_mf_bound}, we have,
\begin{equation}
\begin{small}
\begin{aligned}
\mathbb{E}_{v_i}  \left[ \left\|  \vec{y}_i- \mathbf{B}^{*} f_{\mathrm{gcl}}^{*}(v_i)\right\|_{2}^{2} \right]
\leq \frac{\phi^y }{\hat{\lambda}_{K+1}}
\end{aligned}
\end{small}
\end{equation}
where $\hat{\lambda}_i$ is the $i$-th smallest eigenvalue of the Laplacian matrix $\widehat{\mathbf{L}}_{sym} = \mathbf{I} - \widehat{\mathbf{A}}_{sym}$. Note that $\phi^y$ in Lemma \ref{thm:orig_mf_bound} equals $1- h_{edge}$.  
Then we apply Theorem \ref{thm:homo_sim} and conclude the proof:
\begin{equation}
\begin{small}
\begin{aligned}
\mathbb{E}_{v_i}  \left[ \left\|  \vec{y}_i-\mathbf{B}^{*} f_{\mathrm{gcl}}^{*}(v)\right\|_{2}^{2} \right]
\leq \frac{1 - h_{edge} }{\hat{\lambda}_{K+1}}
\le  \frac{\bar \phi+ \sqrt{ \frac{\sigma^2_{\max}(\mathbf{W}^\top \mathbf{W})  \log(2N^2/\delta)}{2D^2 \|\mathbf{x}^2 \|_{\psi_1}} } }{\hat{\lambda}_{K+1}}
\end{aligned}
\end{small}
\end{equation}